%% file: ms.tex
\documentclass[journal]{IEEEtran}

\ifCLASSINFOpdf
\else
\fi

\usepackage{amsmath,amsfonts,amssymb}
\usepackage{graphicx}
\usepackage{setspace}
\usepackage{tocloft}
\usepackage{adjustbox}
\usepackage{subcaption}
\usepackage{xcolor}
\usepackage{caption}
\usepackage{dblfloatfix} 
\usepackage{breqn}

\begin{document}
\bstctlcite{IEEEexample:BSTcontrol}

\title{Uncertainty-Aware Organ Classification for Surgical Data Science Applications in Laparoscopy}

\author{Sara~Moccia,
        Sebastian J. Wirkert,
        Hannes Kenngott,
        Anant S. Vemuri,
        Martin Apitz,
        Benjamin Mayer,
      Elena~De~Momi,~\IEEEmembership{Senior Member,~IEEE},
        Leonardo S. Mattos,~\IEEEmembership{Member,~IEEE},
        and~Lena Maier-Hein
\thanks{S. Moccia is with the Department
of Electronics, Information and Bioengineering (DEIB), Politecnico di Milano, Milan, Italy, with the Department of Advanced Robotics (ADVR), Istituto Italiano di Tecnologia, Genoa, Italy, and with the Department of Computer Assisted Medical Interventions (CAMI), German Cancer Research Center (DKFZ), Heidelberg, Germany.  e-mail: sara.moccia@polimi.it}
\thanks{S. J. Wirkert,  A. S. Vemuri, and L. Maier-Hein are with the Department of Computer Assisted Medical Interventions (CAMI), German Cancer Research Center (DKFZ), Heidelberg, Germany.}
\thanks{H. Kenngott,  M. Apitz, and B. Mayer are with the Department for General, Visceral, and Transplantation Surgery, Heidelberg University Hospital, Heidelberg, Germany.}
\thanks{E. De Momi is with the Department
of Electronics, Information and Bioengineering (DEIB), Politecnico di Milano, Milan, Italy.}
\thanks{L. S. Mattos is with the Department of Advanced Robotics (ADVR), Istituto Italiano di Tecnologia, Genoa, Italy.}
\\
\thanks{Copyright (c) 2017 IEEE. Personal use of this material is permitted. However, permission to use this material for any other purposes must be obtained from the IEEE by sending an email to pubs-permissions@ieee.org.}
}

%

\maketitle

\begin{abstract}
\textcolor{black}{
Objective: Surgical data science is evolving into a research field that aims to observe everything occurring within
and around the treatment process to provide situation-aware data-driven assistance. In the context of endoscopic video
analysis, the accurate classification
of organs in the field of view of the camera proffers a technical challenge. Herein, we propose a new approach to anatomical structure classification and image tagging that features an intrinsic measure of confidence to estimate its own
performance with high reliability and which can be applied to both RGB and multispectral imaging (MI) data. 
Methods: Organ recognition is performed
using a superpixel classification strategy based on textural and
reflectance information. Classification confidence is estimated
by analyzing the dispersion of class probabilities. Assessment of the proposed
technology is performed through a comprehensive \textit{in vivo} study
with seven pigs.
Results: When applied to
image tagging, mean accuracy in our experiments increased from
65\% (RGB) and 80\% (MI) to 90\% (RGB) and 96\% (MI) with the confidence measure.
Conclusion: Results showed that the confidence measure had a significant influence on the classification accuracy, and MI data are better suited for anatomical structure labeling than RGB data. Significance: This work significantly enhances the state of art in automatic labeling of endoscopic videos by introducing the use of the confidence metric, and by being the first study to use MI data for \textit{in vivo} laparoscopic tissue
classification. The data of our experiments
will be released as the first \textit{in vivo} MI dataset upon
publication of this paper.
}

\end{abstract}


%
\IEEEpeerreviewmaketitle

\section{Introduction}
\label{sec:intro}

Surgical Data Science (SDS) has recently emerged as a new scientific field which aims to improve the quality of interventional healthcare~\cite{maier2017surgical}. 
SDS involves the observation of all elements occurring within and around the treatment process in order to provide the right assistance to the right person at the right time.

In laparoscopy, some of the major opportunities that SDS offers to improve surgical outcomes are surgical decision support~\cite{marz2015toward} and context awareness~\cite{katic2016bridging}. 
Here, technical challenges include the detection and localization of anatomical structures and surgical instrumentation, intra-operative registration, and workflow modeling and recognition. To date, however, clinical translation of the developed methodology continues to be hampered by the poor robustness of the existing methods. In fact, a grand international initiative on SDS \cite{maier2017surgical} concluded that the robustness and reliability of SDS methods are of crucial importance.
With the same perspective, several researches in the case-base reasoning community (e.g.~\cite{cheetham2004measures,kolodner2014case,kendall2017uncertainties}) have pointed out the benefits of estimating method confidence level in assigning a result.
The aim of this paper is to address this issue in the specific context of organ classification and image tagging in endoscopic video images.

\input{tables_images/wf.tex}

Guided by the hypotheses that ({\bf{H1}}) automatic confidence estimation can significantly increase the accuracy and robustness of automatic image labeling methods, and that ({\bf{H2}}) multispectral imaging (MI) data are more suitable for \textit{in vivo} anatomical structure labeling than RGB data, the contributions of this paper are summarized as follows:
\begin{enumerate}
\item {\underline{Uncertainty-aware organ classification}} (Sec.~\ref{sec:uncertainty-aware}): Development of a new method for superpixel ($Spx$)-based anatomical structure classification, which features an intrinsic confidence measure for self-performance estimation and which can be generalized to MI data;
\item {\underline{Automatic image tagging}} (Sec.~\ref{sec:image-tagging}): Development of an approach to automatic image tagging, which relies on the classification method and corresponding confidence estimation to label endoscopic RGB/multispectral images with the organs present in that image;
\item {\underline{\textit{In vivo} validation}} (Sec.~\ref{sec:experiments}): A comprehensive \textit{in vivo} study is conducted using seven pigs to experimentally investigate hypotheses {\bf{H1}} and {\bf{H2}}. 
\end{enumerate}

It is worth noting that, when we mention image tagging, we refer to the action of identifying organs present in an image. Instead, when mentioning organ classification, we refer to the classification of the organ  present in an $Spx$.

To the best of our knowledge, we are the first to use MI data for \textit{in vivo} abdominal tissue classification. Furthermore, this is the first study to address the topic of classification uncertainty estimation. We will make our validation dataset fully available online.

\subsection{Related work}
First attempts at image-guided classification of tissues in RGB endoscopic images primarily used parameter-sensitive morphological operations and intensity-based thresholding techniques, which are not compatible with the high levels of inter-patient multi-organ variability (e.g.~\cite{lee2007automatic, mewes2011automatic}). 
The method for multiple-organ segmentation in laparoscopy reported in~\cite{nosrati2014efficient} relied on non-rigid registration and deformation of pre-operative tissue models on laparoscopic images using color cues. This deformation was achieved using statistical deformable models, which may not always represent the patient-specific tissue deformation, thus resulting in a lack of robustness in terms of inter-patient variability. 
Recently, machine learning based classification algorithms for tissue classification have been proposed to attenuate this issue.
The method described in~\cite{chhatkuli2014live} exploited a machine learning approach to segment the uterus. Gabor filtering and intensity-based features were exploited to segment the uterus from background tissues with support vector machines (SVM) and morphology operators.
However, this approach is limited to single organ segmentation and the performance is influenced by the position of the uterus. 
Similarly, the method presented in~\cite{prokopetc2015automatic} was specifically designed for segmentation of fallopian tubes, as it exploits tube-specific geometrical features, such as orientation and width, and cannot be transferred to other anatomical targets.

In parallel to the development of new computer-assisted strategies to tissue classification, the biomedical imaging field is also evolving thanks to new technologies such as MI~\cite{li2013review}.
MI is an optical technique that enables us to capture both spatial and spectral information on structures. MI provides images that generally have dozens of channels, each corresponding to the reflection of light within a certain wavelength band. Multispectral bands are usually optimized to encode the informative content which is relevant for a specific application.
Thus, MI can potentially reveal tissue-specific optical characteristics better than standard RGB imaging systems \cite{li2013review}.

One of the first \textit{in vivo} applications of MI was proposed by Afromowitz et al. \cite{afromowitz1988multispectral}, who developed a MI system to evaluate the depth of burns on the skin, showing that MI provides more accurate results than standard RGB imaging for such application.
For abdominal tissue classification, Akbari et al.~\cite{akbari2009hyperspectral} and Triana et al.~\cite{triana2014multispectral} exploited pixel-based reflectance features in open surgery and \textit{ex vivo} tissue classification.
The work that is most similar to the present study 
was recently presented by Zhang et al.~\cite{zhang2016tissue}. It pointed out the advantages of combining both reflectance and textural features. However, 
the validation study for this focused on patch-based classification and was limited to \textit{ex vivo} experiments in a controlled environment, including only 9 discrete endoscope poses to view the tissues, with only single organs in the image and without tissue motion and deformation. Furthermore, the challenges of
confidence estimation were not addressed.

As for automatic laparoscopic image tagging, there is no previous work in the literature that has specifically addressed this challenging topic. However, it has been pointed out that there is a pressing need to develop methods for tagging images with semantic descriptors, e.g. for decision support or context awareness \cite{gur2017towards,bodenstedt2016superpixel}. 
For example, context-aware augmented reality (AR) in surgery is becoming a topic of interest. By knowing the surgical phase, it is possible to adapt the AR to the surgeon's needs. Contributions in the field include~\cite{katic2013context,katic2016bridging}. The AR systems in \cite{katic2013context,katic2016bridging} provide context awareness by identifying surgical phases based on (i) surgical activity, (ii) instruments and (iii) anatomical structures in the image. This is something that is commonly assumed as standard~\cite{neumuth2009validation}. However, a strategy for retrieving the anatomical structures present in the image was not proposed.

A possible reason for such a lack in the literature can be seen in the challenging nature of tagging images recorded during \textit{in vivo} laparoscopy. Tissues may look very different across images and may be only partially visible. The high level of endoscopic image noise, the wide range of illumination and the variation of the endoscope pose with respect to the recorded tissues further increase the complexity of the problem. As a result, standard RGB systems may be not powerful enough to achieve the task, even when exploiting advanced machine learning approaches to process the images.
With {\bf{H1}} and {\bf{H2}}, we aim at investigating if the use of MI and the introduction of a measure of classification confidence may face such complexity.

\section{Methods}
\label{sec:met}

Figure~\ref{fig:wf} shows an overview of the workflow of the proposed methods for uncertainty-aware organ classification (Sec.~\ref{sec:uncertainty-aware}) and automatic image tagging (Sec.~\ref{sec:image-tagging}). 
{Table \ref{tab:acron} lists the symbols used in Sec. \ref{sec:met}.}

\begin{table}[tbp]
\caption{{Table of symbols used in Sec. II.}}
\label{tab:acron}
\begin{adjustbox}{width=.5\textwidth}
\begin{tabular}[tbp]{c|l}
Symbol & Description \\
\hline
$N_c$ & Number of image channels\\
$\lambda_i$ & Camera light-filter central wavelength for channel $i$\\
$I(\lambda_i)$ & Row image for channel $i$\\
$Sr(\lambda_i)$ & Spectral reflectance image for channel $i$\\
$D(\lambda_i)$ & Reference dark image for channel $i$\\
$W(\lambda_i)$ & Reference white image for channel $i$\\
$N$ & Number of superpixels in the image\\
$Spx_n$ & $n^{th}$ superpixel $n \in [0,N)$ \\
$LBP_{riu2}^{R,P}$ & Uniform rotation-invariant local binary pattern\\
$R$ & Radius used to compute $LBP_{riu2}^{R,P}$\\
$P$ & Number of points used to compute $LBP_{riu2}^{R,P}$\\
$\{{\mathbf{p}_p\}}_{p \in (0,P-1)}$ & Points used to compute $LBP_{riu2}^{R,P}$\\
$g_{\mathbf{c}}$ & Intensity value of pixel ${\mathbf{c}}$\\
$H_{LBP}$ & Histogram of $LBP_{riu2}^{R,P}$\\
$AS_{Spx_{n}}$ & Average spectrum for $Spx_n$\\
$M$ & Number of pixels in $Spx_n$\\
$l_{H_{LBP}}$ & Length of $H_{LBP}$ for $Spx_n$ and channel $i$\\
$l_{AS}$ & Length of $AS$ for $Spx_n$ and channel $i$\\
$f$ & Support vector machine decision function \\
${\mathbf{x_k}}$ & $k^{th}$ input feature vector \\
$y_k$ & $k^{th}$ output label \\
$\gamma$, $C$ & Support vector machine hyperparameters \\
$N_t$ & Number of training samples\\
$J$ & Total number of considered abdominal tissues\\
$Pr(Spx = j)$ & Probability for the $n^{th}$ $Spx$ to belond to the $j^{th}$ organ\\
$E(Spx_n)$ & Shannon entropy computed for $Spx_n$ \\
$PPCI(Spx_n)$ & Posterior probability certainty index computed for $Spx_n$ \\
$GC(Spx_n)$ & Gini coefficient computed for $Spx_n$ \\
$L$ & Lorentz curve \\
\end{tabular} \\
\end{adjustbox}
\end{table}

\begin{figure*}
\centering
\includegraphics[width = 1\textwidth]{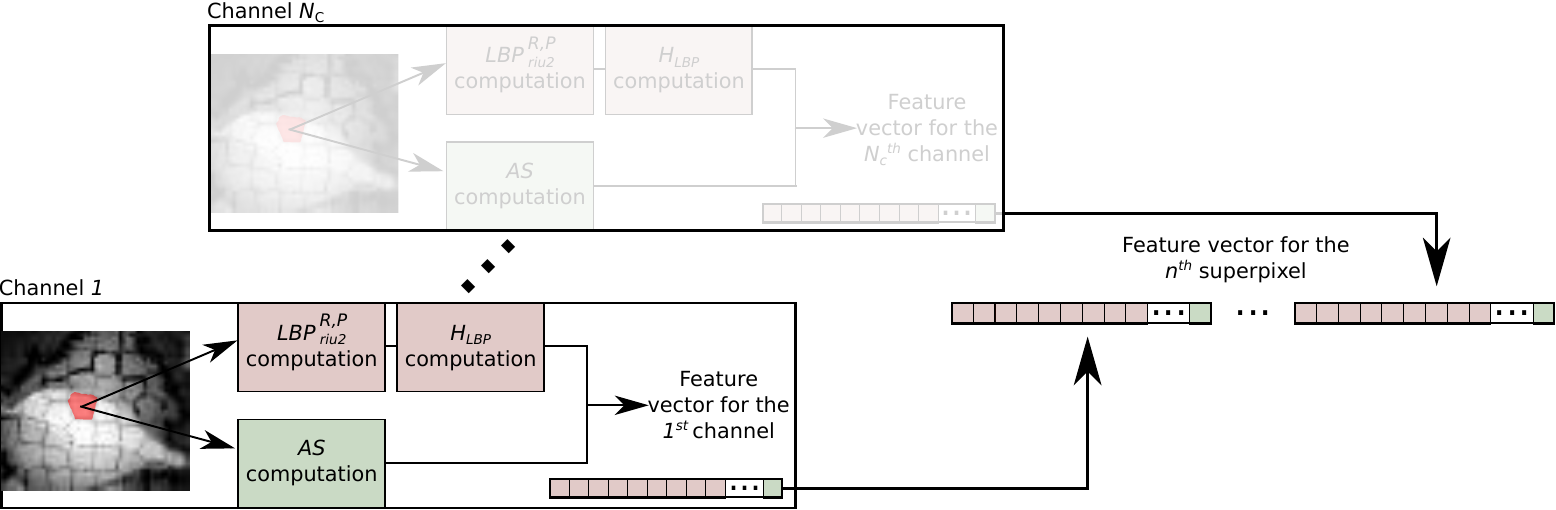}
\caption{\label{fig:spx} A feature vector is extracted from each $n \in N$ superpixel ($Spx_n$), where $N$ is the number of superpixels in the image. The feature vector for $Spx_n$ is obtained by concatenating the histogram ($H_{LBP}$) of uniform rotation--invariant local binary pattern ($LBP_{riu2}^{R,P}$) and the average spectrum ($AS$), for each $i \in N_C$ image channel, where $N_C$ is the number of channels in the image.}
\end{figure*}

\subsection{Uncertainty-aware tissue classification}
\label{sec:uncertainty-aware}
The steps comprising the proposed approach to organ classification are presented in the following subsections.

\subsubsection{Pre-processing}
To remove the influence of the dark current and to obtain the spectral reflectance image $Sr(\lambda_i)$ for each MI channel $i \in [1, N_C]$), where $N_C$ is the number of MI bands, the raw image $I(\lambda_i)$ was pre-processed by subtracting the reference dark image $D(\lambda_i)$ of the corresponding channel from the multispectral image. $\lambda_i$ refers to the band central wavelength of the $i^{th}$ channel. This result was then divided by the difference between the reference white image $W(\lambda_i)$ of the corresponding channel and $D(\lambda_i)$, as suggested in~\cite{mansouri2005development}:

\begin{equation}
\label{eq:norm}
Sr(\lambda_i) = \frac{I(\lambda_i)-D(\lambda_i)}{W(\lambda_i)-D(\lambda_i)}
\end{equation}

Note that $W(\lambda_i)$ and $D(\lambda_i)$ had to be acquired only once for a given camera setup and wavelength. These images were obtained by placing a white reference board in the field of view and by closing the camera shutter, respectively. Each reflectance image was additionally processed with anisotropic diffusion filtering to remove noise while preserving the sharp edges~\cite{kroon2010optimized}. The specular reflections were segmented by converting the RGB image into hue, saturation, value (HSV) color space and thresholding the V value. They were then masked from all channels \cite{moccia2016automatic}. 

\subsubsection{Feature extraction}
In the method proposed in this study, we extracted features from $Spx$. $Spx$ were selected because, compared to regular patches, they are built to adhere to image boundaries better~\cite{li2015superpixel}. This characteristic is particularly useful considering the classification of multiple organs within one single image. 
To obtain the $Spx$ segmentation, we applied linear spectral clustering (LSC)~\cite{li2015superpixel} to the RGB image and then used the obtained $Spx$ segmentation for all multispectral channels.

Inspired by the recently published \textit{ex vivo} study by Zhang et al.~\cite{zhang2016tissue}, we extracted both textural and spectral reflectance features from each multispectral channel. 
Indeed, as stated in Sec.~\ref{sec:intro}, the authors demonstrated that incorporating textural information improved the classification performance with respect to single pixel-based features in their controlled experimental setup.
As  laparoscopic  images  are  captured  from  various  viewpoints  under  various  illumination
conditions,  the  textural  features  should  be  robust  to  the  pose  of  the  endoscope  as  well  as  to  the  lighting conditions.  Furthermore, their computational cost should be negligible to enable real-time computation with a view to future clinical applications.

The histogram ($H_{LBP}$) of the uniform rotation--invariant local binary pattern ($LBP_{riu2}^{R,P}$), which fully meets these requirements, was here used to describe the tissue texture of an $Spx$. 

The $LBP^{R,P}_{riu2}$ formulation requires to define, for a pixel 
$\mathbf{c} = (c_x, c_y)$, a spatial circular neighborhood of radius $R$ with $P$ equally-spaced neighbor points ($\{{\mathbf{p}_p\}}_{p \in (0,P-1)}$):
\begin{equation}
LBP^{R,P}_{riu2}(\mathbf{c}) = \begin{cases} 
								\sum_{p=0}^{P-1}s(g_{{\mathbf{p}}_p} - g_{\mathbf{c}}), & \mbox{if } U(LBP^{R,P}) \leq 2 \\ 
								P+1, & \mbox{otherwise} 
                                \end{cases}
\end{equation}
where $g_{\mathbf{c}}$ and $g_{{\mathbf{p}}_p}$ denote the gray values of the pixel $\mathbf{c}$ and of its $p^{th}$ neighbor $\mathbf{p}_p$, respectively. $s(g_{\mathbf{p}_p} - g_{\mathbf{c}})$ is defined as:
\begin{equation}
s(g_{\mathbf{p}_p} - g_{\mathbf{c}}) = \Bigg\{
\begin{array}{rl}
1, & \text{$g_{\mathbf{p}_p} \geq g_{\mathbf{c}}$}\\
0, & \text{$g_{\mathbf{p}_p} < g_{\mathbf{c}}$}
\end{array}
\end{equation}
and $U(LBP^{R,P})$ is defined as:
\begin{equation}
\begin{split}
U(LBP^{R,P}) = |s(g_{\mathbf{p}_{P-1}}-g_{\mathbf{c}}) -s(g_{\mathbf{p}_0}-g_{\mathbf{c}})| + \\
\sum_{p=1}^{P-1}|s(g_{\mathbf{p}_{p}}-g_{\mathbf{c}}) -s(g_{\mathbf{p}_{p-1}}-g_{\mathbf{c}})|
\end{split}
\end{equation}

The $H_{LBP}$, which counts the occurrences of $LBP^{R,P}_{riu2}$, was normalized to the unit length to account for the different pixel numbers in an $Spx$.

Spectral reflectance information was encoded in the average spectrum $(AS)$, which is the average spectral reflectance value in an $Spx$. 
The $AS$ for the $i^{th}$ channel and the $n^{th}$ $Spx$ ($Spx_n$), with $n \in (1,N)$ and $N$ the total number of $Spx$, is defined as:
\begin{equation}
AS_{Spx_n}(\lambda_i) = \frac{1}{M} \sum_{\mathbf{p} \in Spx_n} Sr_p(\lambda_i)
\end{equation}

where {$M$} is the number of pixels in $Spx_n$ and $Sr_p(\lambda_i)$ is the reflectance value of the $p^{th}$ pixel of $Spx_n$ in the $i^{th}$ channel.

The L2-norm was applied to the $AS$ in order to accommodate lighting differences. 
$AS$ was exploited instead of the simple spectral reflectance at one pixel to improve the feature robustness against noise, although this is detrimental to spatial resolution.

%
The steps for obtaining the feature vector are shown in Fig.~\ref{fig:spx}.

\subsubsection{Superpixel-based classification}
To classify the $Spx$-based features, we used SVM with the radial basis function.
For a binary classification problem, given a training set of $N_t$ data $\{y_k, {\mathbf{x_k}}\}_{k=1}^{N_t}$, where ${\mathbf{x_k}}$ is the $k^{th}$ input feature vector and $y_k$ is the $k^{th}$ output label, the SVM decision function ($f$) takes the form of:
\begin{equation}
f({\mathbf{x}}) = sign\Big[\sum_{k=1}^{N_t} a_k^* y_k \Psi({\mathbf{x}}, {\mathbf{x_k}}) + b \Big]
\end{equation}
where:
\begin{equation}
\Psi({\mathbf{x}}, {\mathbf{x_k}}) = exp\{-\gamma||{\mathbf{x}}-{\mathbf{x_k}}||_2^2/\sigma^2\}, \qquad \gamma > 0
\end{equation}
$b$ is a real constant and $a_k^*$ is computed as follows:
\begin{equation}
a_k^* = \max \Big \{ -\frac{1}{2} \sum_{k,l=1}^{N_t} y_k y_l \Psi({\mathbf{x_k}}, {\mathbf{x_l}}) a_k a_l + \sum_{k=1}^{N_t} a_k \Big \}
\end{equation}
with:
\begin{equation}
\sum_{k=1}^{N_t} a_k y_k = 0, \qquad 0 \leq a_k \leq C, \qquad k=1, ..., {N_t}
\end{equation}
In this paper, $\gamma$ and $C$ were computed with grid search, as explained in Sec. \ref{sec:experiments}.

\begin{figure}[tbp]
\centering
\includegraphics[width = .25\textwidth]{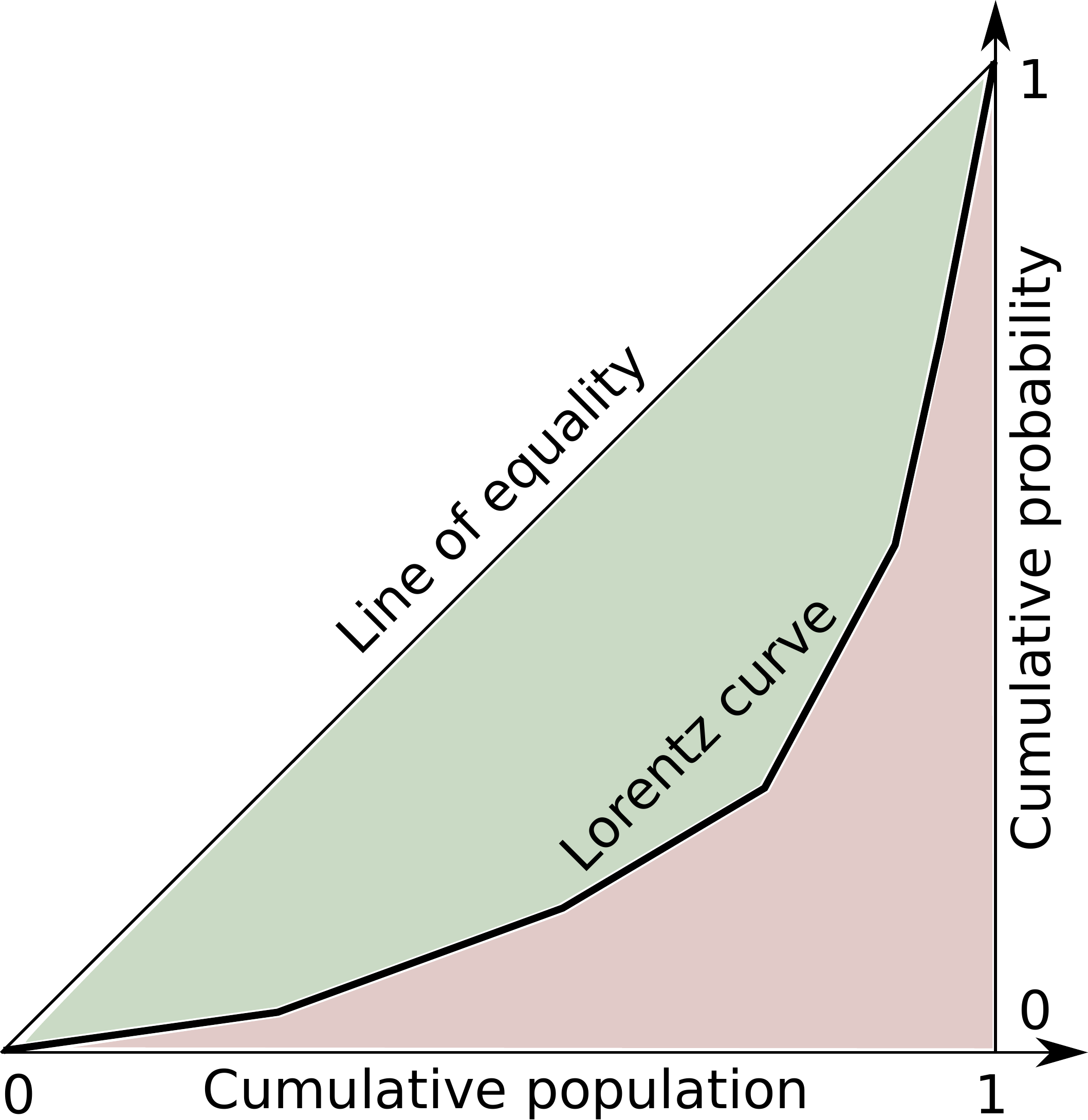}
\caption{The Gini coefficient ($GC$) is computed as twice the area (green area) between the line of equality and the Lorentz curve. The Lorentz curve represents the cumulative classification probability among the outcome classification states rank-ordered according to the decreasing values of their individual probabilities. A uniform discrete probability distribution has $GC=0$, as the Lorentz curve overlays the line of equality, while for a state with probability 100\% and the others at 0\%, $GC=1$.}
\label{fig:gc}
\end{figure}

Since our classification task is a multiclass classification problem, we implemented SVM with the \textit{one-against-one} scheme. Specifically, six organ classes were involved in the SVM training process, as described in Sec. \ref{sec:experiments}.
Prior to classification, we standardized the feature matrix within each feature dimension.

As a prerequisite for our confidence estimation, we retrieved the probability {$Pr(Spx_n=j)$} for the $n^{th}$ $Spx$, to belong to the $j^{th}$ organ ($j\in [1,J]$), $J$ is the number of considered organs. 
In particular, {$Pr(Spx_n=j)$} was obtained, according to the pairwise comparison method proposed in \cite{wu2004probability} (which is an extension of \cite{platt1999probabilistic} for the binary classification case), by solving:
\begin{equation}
Pr(Spx_n = j) = \sum_{i=1,i \neq j}^J \frac{Pr(Spx_n = j) + Pr(Spx_n = i)}{J-1}r_{ji}, \forall j
\end{equation}
subject to:
\begin{equation}
\sum_{j=1}^J Pr(Spx_n = j) = 1, \quad Pr(Spx_n = j) \geq 0, \quad  \forall j
\end{equation}
where $r_{ij}$ is the estimates of $Pr(Spx_n = j| Spx_n \in \{i,j\})$ with $r_{j,i} + r_{i,j} = 1, \forall j \neq i$. The estimator $r_{j,i}$ was obtained according to \cite{platt1999probabilistic}, mapping the SVM output to probabilities by training the parameters of a sigmoid function.

\input{tables_images/camera_spec.tex}

\subsubsection{Confidence estimation}
To estimate the SVM classification performance, we evaluated two intrinsic measures of confidence: (i) a measure based on the normalized Shannon entropy~($E$), called posterior probability certainty index ($PPCI$), and (ii) the Gini coefficient ($GC$)~\cite{marcot2012metrics}.

For the $n^{th}$ $Spx$, $PPCI(Spx_n)$ is defined as:
\begin{equation}
PPCI(Spx_n) = 1 - E(Spx_n)
\end{equation}
where $E$ is:
{
\begin{equation}
E(Spx_n) = -\frac{\sum_{j=1}^J Pr(Spx_n=j) log (Pr(Spx_n=j))}{log(J)}
\end{equation}
and:
\begin{equation}
\!\begin{aligned}[l]
log (Pr(Spx_n=j)) = \\
\begin{cases} 
log(Pr(Spx_n=j)), & \mbox{if } Pr(Spx_n=j)>0 \\ 
0, & \mbox{if } Pr(Spx_n=j)=0 
\end{cases}
\end{aligned}
\end{equation}
}

For the $n^{th}$ $Spx$, $GC(Spx_n)$ is defined as:
\begin{equation}
\label{eq:gc}
GC(Spx_n) = 1 - 2 \int_0^1 \! L(x) \, \mathrm{d}x. 
\end{equation}
where $L$ is the Lorentz curve, which is the cumulative probability among the $J$ outcome states rank-ordered according to the decreasing values of their individual probabilities 
{($Pr(Spx_n = 1), ..., Pr(Spx_n = J)$)}.
As can be seen from Fig.~\ref{fig:gc}, in case of uniform discrete probability distribution (complete uncertainty), $L$ corresponds to the line of equality. Thus, the integral in Eq. \ref{eq:gc} (red area in Fig.~\ref{fig:gc}) has values 0.5 and $GC = 0$. On the contrary, for the case of a single state at 100\% with the others at 0\% (complete certainty), the integral value is 0 and $GC=1$. The $GC$ computation can be also seen as twice the area (green area in Fig.~\ref{fig:gc}) between the line of equality and the Lorentz curve.

\begin{figure*}
\centering
\includegraphics[width = .6\textwidth]{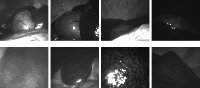}
\caption{Challenges of the evaluation dataset. 
Four samples of images showing the gallbladder (first row) and spleen (second row) are reported. Images were recorded with varying endoscope pose and illumination level. Specular reflections are present in the images due to the smooth and wet organ surfaces. Multiple organs can be present in a single image. All images refer to the same multispectral channel.}
\label{fig:chall}
\end{figure*}

Although both metrics are suitable to evaluate the dispersion of the classification probability, $GC$ is faster to compute, as it does not require the logarithm computation. Moreover, $GC$ is more sensitive than $PPCI$ at higher values~\cite{marcot2012metrics}.



\subsection{Automatic image tagging}
\label{sec:image-tagging}
Automatic image tagging uses the SVM $Spx$-based classification and the corresponding confidence estimation.
Specifically, test images were tagged considering $Spx$ labels with high confidence values only. 
The value of $GC(Spx_n)$ was thresholded to obtain binary confidence information. An $Spx$ was considered to have an acceptable confidence level if $GC(Spx_n) > \tau$, for the threshold~$\tau$. 
The same procedure was performed using $PPCI$ instead of $GC$.

\input{tables_images/feat_conf.tex}

\section{In vivo validation} 
\label{sec:experiments}
Seven pigs were used to examine the {\bf{H1}} and {\bf{H2}} introduced in Sec.~\ref{sec:intro}. 
Raw multispectral images ($I$) were acquired using a custom-built MI laparoscope. 
In this study, the multispectral laparoscope was comprised of a Richard Wolf (Knittlingen, Germany) laparoscope and a 5--MP Pixelteq Spectrocam (Largo, FL, USA) multispectral camera.
The $\lambda_i$ for each $i^{th}$ band index and the corresponding full widths at half maximum (FWHM) are reported in Table~\ref{tab:wave}.
The filters were chosen according to the band selection strategy for endoscopic spectral imaging presented in \cite{wirkert2014endoscopic}. The method makes use of the Sheffield index \cite{sheffield1985selecting},  which is an information theory based band selection method originally proposed by the remote sensing community.
The $700$, $560$ and $470$ nm channels were chosen to simulate RGB images as the camera did not provide RGB images directly.  
The image size was $1228\times1029\times8$ for MI and $1228\times1029\times3$ for RGB.

The physical size of the multispectral camera was 136~x~124~x~105~mm, with a weight of 908~g. The acquisition time of one multispectral image stack took 400~ms.

From the seven pigs, three pigs were used for training ($29$ images) and four for testing ($28$ images). 
The number of images used to test the SVM performance on RGB and MI data was the same, as RGB data were directly obtained from MI data by selecting 3 of the 8 MI channels. The total number of $Spx$ in the training and testing dataset, for both MI and RGB data, was 1382 and 1559, respectively.


We considered six porcine organ tissues typically encountered during hepatic laparoscopic surgery: the liver, gallbladder, spleen, diaphragm, intestine, and abdominal wall. These tissues were recorded during \textit{in vivo} laparoscopy. 
Challenges associated with the \textit{in vivo} dataset include:
\begin{itemize}
\item Wide range of illumination
\item
Variation of the endoscope pose
\item Presence of specular reflections
\item Presence of multiple organs in one image
\item Organ movement
\end{itemize}
Visual samples of the dataset challenges are shown in Fig. \ref{fig:chall}.

The multispectral images were pre-processed as described in Sec.~\ref{sec:uncertainty-aware}. 
The $Spx$ segmentation with LSC was achieved using an average $Spx$ size of $150^2$ pixels and an $Spx$ compactness factor of $0.1$. Accordingly, $55$ $Spx$ on average were obtained for each image.
The $LBP_{riu2}^{R,P}$ were computed considering the following $(R, P)$ combinations: {(1, 8), (2, 16), and (3,~24)}.
The feature vector for an $Spx$ was obtained by concatenating the $H_{LBP}$ with the $AS$ value for all $8$ multispectral channels (for MI) and for $\lambda_i = 700$, $560$ and $470$ nm (for RGB).  
The feature vector size for an $Spx$ was: 
\begin{equation}
(l_{H_{LBP}} + l_{AS}) \times N_C
\end{equation}
where $l_{H_{LBP}}$ is the length of $H_{LBP}$, equal to 54, $l_{AS}$ is the length of $AS$, equal to 1, and $N_C$ is the number of channels, 3 for RGB and 8 for multispectral data.

The SVM kernel parameters ($C = 10^4$ and $\gamma=10^{-5}$) were retrieved during the training phase via grid-search and 10-fold cross-validation on the training set. The grid-search spaces for $\gamma$ and $C$ were set to [$10^{-8}$,~$10^{1}$] and [$10^{1}$, $10^{10}$], respectively, with $10$ values spaced evenly on the $log_{10}$ scale in both cases.
The determined values for the hyperparameters were subsequently used in the testing phase. 

The feature extraction was implemented using OpenCV \footnote{http://opencv.org/}.
The classification was implemented using scikit-learn \cite{pedregosa2011scikit} \footnote{http://scikit-learn.org/}.

\input{tables_images/spAccThConf.tex}

\subsubsection{Investigation of H1}
To investigate whether the inclusion of a confidence measure increases $Spx$-based organ classification accuracy ($Acc_{Spx}$), we evaluated the $Acc_{Spx}$ dependence on $\tau \in [0.5:0.1:1)$ applied to both $GC$ and $PPCI$.
$Acc_{Spx}$ is defined as the ratio of correctly classified confident $Spx$ to all confident samples in the testing set.
We evaluated whether differences existed between $Acc_{Spx}$ obtained applying $GC$ and $PPCI$ on the SVM output probabilities using the Wilcoxon signed-rank test for paired samples (significance level = 0.05).
{We also investigated the SVM performance with the inclusion of confidence when leaving one organ out of the training set. Specifically, we trained six SVMs, leaving each time one organ out. We computed, for each of the six cases, the percentage ($^\%LC_{Spx}$) of low-confidence $Spx$ (considering $\tau = 0.9$). We did this both for the organ that was excluded ($Ex$) from the training set and for the included organs ($In$).}
For image tagging, we computed the tagging accuracy ($Acc_{Tag}$) for different $\tau$, where $Acc_{Tag}$ is the ratio of correctly classified organs in the image to all organs in the testing image.

\subsubsection{Investigation of H2}
To investigate whether MI data are more suitable for anatomic structure classification than conventional RGB video data, we performed the same analysis for RGB and compared the results with those from the MI. 
To complete our evaluation, we also evaluated the performance of $H_{LBP}$ alone and $AS$ alone for $\tau=0$, which corresponds to the \textit{Base} case, i.e., SVM classification without a confidence computation.
Since the analyzed populations were not normal, we used the Wilcoxon signed-rank test for paired samples to assess whether differences existed between the mean ranks of the RGB and MI results (significance level~$= 0.05$). 

\input{tables_images/cm.tex}

\input{tables_images/hl_acc1.tex}

\section{Results}
\label{sec:res}

The descriptive statistics of $Acc_{Spx}$ for the analyzed features are reported in Table~\ref{tab:feat_conf}.
For the \textit{Base} case, the highest $Acc_{Spx}$ (median $= 90\%$, inter-quartile range $= 6\%$) was obtained with $H_{LBP}+AS$ and MI.
The other results all differ significantly (p-value $< 0.05$) from those obtained with $H_{LBP}+AS$ and MI.

When $\tau$ applied to $GC$ (Fig. \ref{fig:spAccThConf_gc}) and $PPCI$ (Fig. \ref{fig:spAccThConf_entro}) was varied in [0.5 : 0.1 : 1), the median $Acc_{Spx}$ for the MI data increased monotonously to 99\% ($\tau = 0.9$), when using both $GC$ and $PPCI$. The same trend was observed for the RGB data, with an overall improvement of the median from 81\% to 93\% (using $GC$) and 91\% (using $PPCI$). 
For both the \textit{Base} case and after introduction of the confidence measures, the MI outperformed the RGB (p-value $<$ 0.05). No significant differences were found when comparing the classification performance obtained with $GC$ and $PPCI$. Therefore, as $GC$ computation is more sensitive to high values and faster to compute than $PPCI$, we decided to use $GC$.

\begin{figure}
\centering
\includegraphics[width = .4\textwidth]{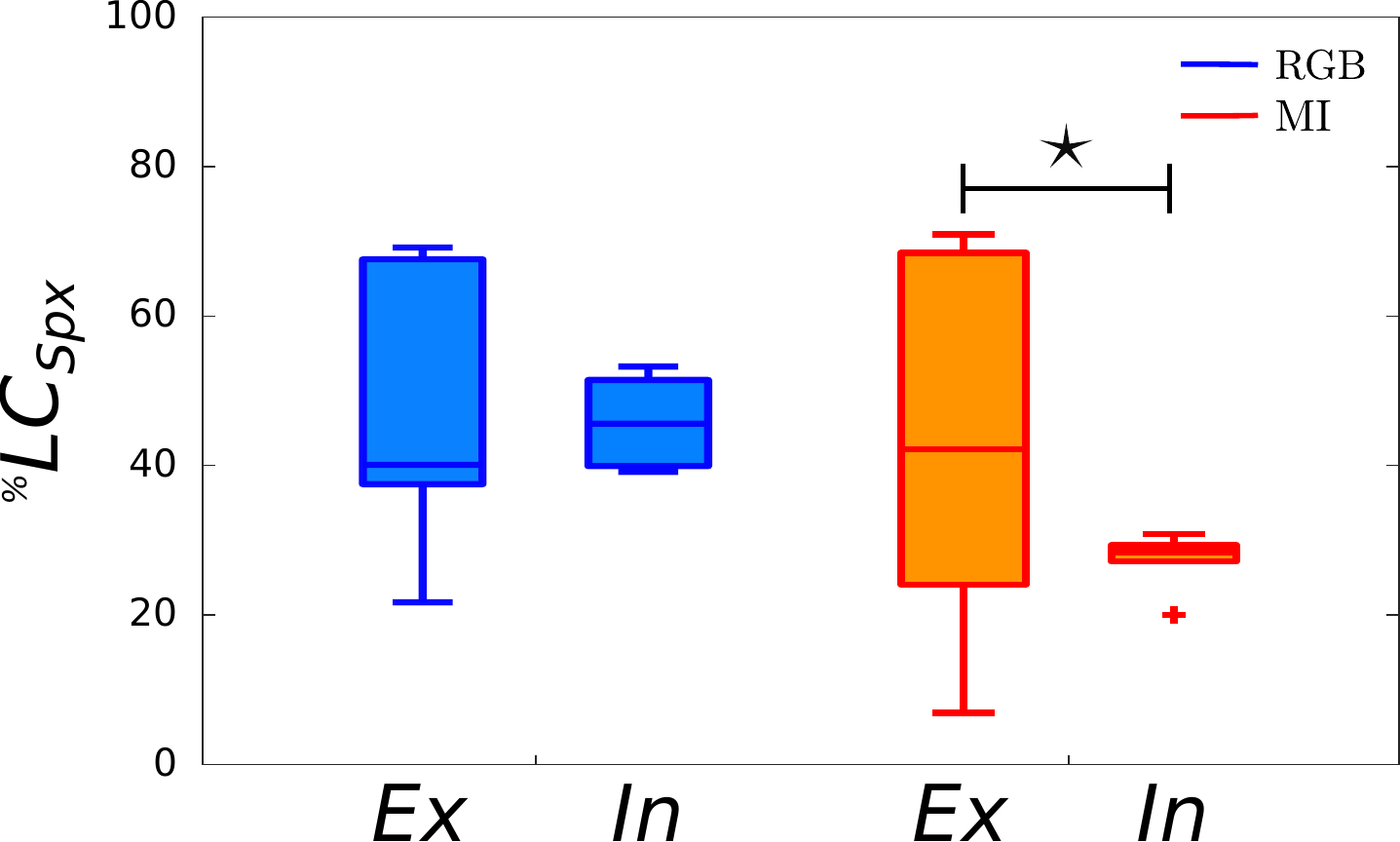}
\caption{\label{fig:in_ex} {Effect of previously unseen target structures on the uncertainty estimation. Percentage ($^\%LC_{Spx}$) of low-confidence $Spx$ ($\tau = 0.9$) for organs that were seen ($In$) and not seen ($Ex$) in the training process.}}
\end{figure}

Figure~\ref{fig:cm} shows the confusion matrix for MI and $\tau = 0.9$ on $GC$. Note that, in the case yielding the least accurate result, which corresponds to spleen classification, the accuracy rate still achieved $96\%$, whereas for RGB the lowest accuracy rate was~$69\%$.

\input{tables_images/tag_base_conf.tex}

\input{tables_images/visual.tex}

The $^\%LC_{Spx}$ boxplots relative to the leave-one-organ out experiment are shown in Fig. \ref{fig:in_ex}. The $^\%LC_{Spx}$ is significantly higher for organs that were not seen in the training phase (MI: 42\% ($Ex$) vs.  23\% ($In$); RGB: 36\% ($Ex$) vs. 40\% ($In$)).

When applied to endoscopic image tagging, the mean $Acc_{Tag}$ values in our experiments were increased from 65\% (RGB) and 80\% (MI) to 90\% (RGB) and 96\% (MI) with the incorporation of the confidence measure (using $GC$). The descriptive statistics are reported in Fig.~\ref{fig:hl_acc1}. In this instance, the MI also outperformed the RGB both in the \textit{Base} case and with the confidence measure (p-value~$<0.05$). {Figure~\ref{fig:tag_base_conf}} shows the influence of low-confidence $Spx$ exclusion on the image tagging: after low-confidence $Spx$ exclusion, all $Spx$ in the image were classified correctly.

Sample results for the SVM classification and the corresponding confidence map (using $GC$) are shown in {Fig.~\ref{fig:visual}}. For low-confidence $Spx$, the probable cause of uncertainty is also reported. The main sources of uncertainty are specular reflections, camera sensor noise at the image corner, and the partial organ effect, i.e., when two or more organs correspond to one $Spx$. 

\section{Discussion}
\label{sec:dis}

The emerging field of surgical data science \cite{maier2017surgical} aims at observing the entire patient workflow in order to provide the right assistance at the right time. One important prerequisite for context-aware assistance during surgical treatment is to correctly classify the phase within an intervention. While a great amount of effort has been put into automatic instrument detection (e.g. \cite{sznitman2014fast,bodenstedt2016superpixel,allan2013toward}), the problem of automatic organ classification  has received extremely little extension. We attribute this to the fact that the task is extremely challenging. In fact, the related problem of organ boundary detection was regarded so challenging by participants of the MICCAI 2017 endoscopic vision challenge (https://endovis.grand-challenge.org/) that only a single team decided to submit results for the sub-challenge deadline with kidney boundary detection. In this work, we tackled this problem by two previously unexplored approaches:
\begin{itemize}
\item Accuracy: We slightly changed the image acquisition process using a multispectral camera as opposed to a standard RGB camera in order to increase the quality of the input data (for the classifier). The effect of this measure was an increase in accuracy of 11\% for the task of organ classification and an increase of 23\% for the task of automatic image tagging. 
\item Robustness: We derived superpixel-based measures of confidence to increase the reliability of image tagging. The result was a boost in accuracy of 38\% (RGB) and 20\% (MI) absolute.
\end{itemize}

With our validation dataset, we showed that  MI significantly outperforms standard RGB imaging in classifying abdominal tissues.
Indeed, as the absorption and scattering of light in tissue is highly dependent on (i) the molecules present in the tissues, and (ii) the wavelength of the light, 
the multispectral image stack was able to encode the tissue-specific optical information, enabling higher accuracy in distinguish different abdominal structures in comparison to standard RGB.


With the introduction of the confidence measure, 
we showed that the classification accuracy can be improved, for both  RGB and MI.
This happened when exploiting  both $GC$ and $PPCI$. Since no significant differences were found between $GC$ and $PPCI$, we decided to use $GC$ as it is more sensitive at higher values than $PPCI$ and its computation is faster.
In fact, a major advantage of our method is its high classification accuracy, which attained 93\% (RGB) and 99\% (MI) in the regions with high confidence levels, with a significant improvement compared to the \textit{Base} case.
Few misclassifications of high-confidence $Spx$ occurred, and where they did then this was mainly with tissues that are also challenging to distinguish between for the human eye, e.g. liver and spleen
(Fig.~\ref{fig:cm}). 
{
It is worth noting that $GC$ and $PPCI$ were two examples of confidence estimation measures {to investigate {\bf{H1}}}. {We decided against using simple thresholding on the maximum ($Max$) value of $Pr(Spx_n=j)$ computed among the $J$ organ classes as $GC$ and $PPCI$ are generally known for being more sensitive at higher values \cite{marcot2012metrics}. This assumption was confirmed in additional experiments, where image tagging performed with confident $Spx$ according to $GC$/$PPCI$ was substantially more robust than tagging based on confident $Spx$ according to $Max$.}

The results obtained with the introduction of the confidence measure are comparable with those obtained by Zhang et al.~\cite{zhang2016tissue} for \textit{ex vivo} organ classification in a controlled experimental setup. Zhang et al. reported a median classification accuracy of 98\% for MI, whereas our classification accuracy for the \textit{Base} case only achieved 90\% due to the challenging nature of the \textit{in vivo} dataset.
An accuracy level comparable to the one of~\cite{zhang2016tissue} was, however, restored for our dataset once the low-confidence $Spx$ were excluded.

{When excluding one organ from the training set,  $^\%LC_{Spx}$ relative to the excluded organ was significantly higher than the number of low-confidence superpixels obtained for the remaining organs. This indicates that the confidence inclusion helped in handling situations where unknown structures appeared in the field of view of the camera.}

These results are in keeping with those found in the literature for case reasoning~\cite{cheetham2004measures, kendall2017uncertainties}. Indeed, the importance of the estimation of the level of confidence of the classification with a view to improving system performance has been widely highlighted in several research fields, such as face recognition~\cite{delany2005generating}, spam-filtering~\cite{ orozco2008confidence}, and cancer recognition~\cite{zhang2013subpopulation}.
However, the use of confidence metrics had not been exploited in the context of laparoscopic image analysis, up until now.

Although several $Spx$ misclassifications occurred at the \textit{Base} case, which had a negative effect on tagging performance, the low-confidence $Spx$ exclusion significantly increased tagging accuracy.
Indeed, regions affected by camera sensor noise, specular reflections, and spectral channel shift due to organ movement were easily discarded based on their confidence value. The same process was implemented when the $Spx$ segmentation failed to separate two organs.
Also in this case, MI showed that it performs better than standard RGB.


While we are the first to address the challenges of \textit{in vivo} image labeling, including the large variability of illumination, variation of the endoscope pose, the presence of specular reflections, organ movement, and the appearance of multiple organs in one image, one disadvantage of our validation setup is that our database was not recorded during real surgery. Hence, some of the challenges typically encountered when managing real surgery images were absent (e.g., blood, smoke, and occlusion). 
Moreover, as our camera does not provide RGB data directly, we generated a synthetic RGB image by merging three MI channels. It should be noted, however, that our RGB encodes more specific information, as the bands used to obtain these data are considerably narrower than those of standard RGB systems (FWHM = $20$ nm). 
We also recognize that a limitation of the proposed work could be seen in the relatively small number of training images (29). 
However, analyzing researches on the topic of tissue classification in laparoscopy, such number is comparable with the one of Chhatkuli et al. \cite{chhatkuli2014live}, which exploited 45 uterine images, and Zhang et al. \cite{zhang2016tissue}, which recorded 9 poses of just 12 scenes (3 pigs $\times$ 4 \textit{ex-vivo} organs). Further, it is worth noting that our training was performed at $Spx$-level, meaning that the training set sample size was about $55 \times 29$, where 55 is the average number of $Spx$ in an image.

Considering that the proposed study was not aimed at evaluating the system performance for clinical translation purpose, we did not analyze the clinical requirements of the proposed method performance. Despite the fact that we recognize the relevance of such analysis, we believe that it should be performed in relation to the specific application. 
For example, with reference to \cite{katic2013context}, we plan to analyze and evaluate the requirements of a context-aware AR system supported by the proposed methodology. 
{
However, when discussing with our clinical partners, it emerged that the end-to-end accuracy should be close to 100\% (i.e. for recognizing the surgical state). However, it has to be further investigated how errors in image tagging affect the error of the final task.}

With our MI laparoscope prototype, the image stack acquisition time (400 ms) was faster than most systems commonly presented in literature, like e.g. (e.g. \cite{clancy2014multispectral} with $\sim$3~s), which makes it more advantageous for clinical applications. Anyway, to fully meet the clinical requirements in terms of system usability, we are currently working on further shrinking the system and speeding it up, as to achieve real-time acquisition.
{A further solution we would like to investigate is the use of loopy belief propagation \cite{murphy1999loopy,ihler2005loopy} as post-processing strategy to include spatial information with respect to how confident classification labels appear in the image. 
This would be particularly useful for images where the tagging failed due to few confident misclassified $Spx$ surrounded by correctly classified confident $Spx$.}
Future work will also deal with the real-time implementation of the classification algorithm, which was not the aim of this work. Recent advancements in tissue classification research suggest that the use of convolutional neural network (CNN) could be also investigated for comparison~\cite{shin2016deep}. 
{Indeed, uncertainty in deep learning is an active and relatively new field of research, and standard deep learning tools for classification do not capture model uncertainty \cite{gal2016uncertainty}. Excluding popular dropout strategies (e.g. \cite{srivastava2014dropout,kendall2016modelling}), among the most recently proposed solutions, variational Bayes by Backpropagation \cite{blundell2015weight,pawlowski2017implicit} is is drawing attention of the deep learning community.}

\section{Conclusions}
\label{sec:con}

In this paper, we addressed the challenging topic of robust classification of anatomical structures in \textit{in vivo} laparoscopic images. 
With the first \textit{in vivo} laparoscopic MI dataset, we confirmed the two hypotheses: ({\bf{H1}}) the inclusion of a confidence measure increases the $Spx$-based organ classification accuracy substantially and ({\bf{H2}}) MI data are more suitable for anatomic structure classification than conventional video data.
To this end, we proposed the first approach to anatomic structure labeling. The approach features an intrinsic confidence measure and can be used for high accuracy image tagging, with an accuracy of $90\%$ for RGB and $96\%$ for MI. 
In conclusion, the method proposed herein could become a valuable tool for surgical data science applications in laparoscopy due to the high level of accuracy it provides in image tagging. Moreover, by making our MI dataset fully available, we believe we will stimulate researches in the field, encouraging and promoting the clinical translation of MI systems.

\section*{Acknowledgments}
The authors would like to acknowledge support from the European Union through the
ERC starting grant COMBIOSCOPY under the New Horizon Framework Programme
grant agreement ERC-2015-StG-37960.


\subsection*{Compliance with ethical standards}

\subsection*{Disclosures}
The authors have no conflict of interest to disclose.
\subsection*{Ethical standards} 
This article does not contain any studies with human participants. All applicable international, national and/or institutional guidelines for the care and use of animals were followed.


\bibliographystyle{IEEEtran}  
\bibliography{bibtex/bib/biblio.bib}   

\end{document}

%% file: tables_images/wf.tex
\begin{figure*}[tbp]
    \centering
        \includegraphics[width=\textwidth]{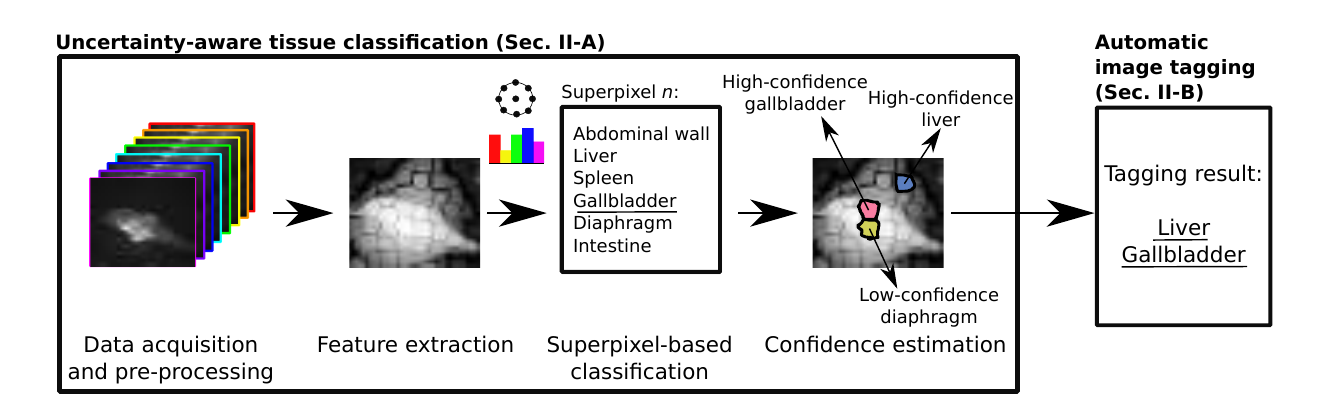}
        \caption{Workflow of proposed approaches for uncertainty-aware organ classification and automatic image tagging.\\}
        \label{fig:wf}
\end{figure*}

%% file: tables_images/camera_spec.tex
\begin{table*}[tbp]
\centering
\caption{Camera light-filter central wavelengths and full width at half maximum (FWHM) for each $i (=1$--$8)$ band in multispectral imaging (MI) and RGB.}
\label{tab:wave} 
\begin{adjustbox}{max width=\textwidth}
\begin{tabular}[tbp]{ccccccccc}
\hline
$i$		& $1$		& $2$ 		&$3$			& $4$ 		& $5$		&  $6$		& $7$		& $8$	 \\
\hline
MI 		& $470$ nm	& $480$ nm 	& $511$ nm	& $560$ nm 	& $580$ nm 	&  $600$ nm 	& $660$ nm 	& $700$ nm 	\\
RGB 	& $470$ nm	& -		 	& -		 	& $560$ nm 	& - 			&  - 			& -			 & $700$ nm 	\\
FWHM	& $20$ nm		& $25$ nm 	&$20$ nm		& $20$ nm 	& $20$ nm		&  $20$ nm	& $20$ nm		 & $20$ nm	 \\		
\hline
\end{tabular}
\end{adjustbox}
\end{table*}

%% file: tables_images/feat_conf.tex
\begin{table*}[tbp]
\centering
\caption{Median superpixel-based accuracy rate ($Acc_{Spx}$) and inter-quartile range (in brackets) for RGB and multispectral imaging (MI) using different features for the \textit{Base} case (i.e., without confidence inclusion). $H_{LBP}$: Histogram of local binary patterns; $AS$: Average spectrum.}
\label{tab:feat_conf} 
\setlength\tabcolsep{6pt}
\begin{adjustbox}{max width=\textwidth}
\begin{tabular}[tbp]{cccccccccc}
\hline

		&\multicolumn{2}{c}{$H_{LBP}$}		&	\multicolumn{2}{c}{ $AS$}			&	\multicolumn{2}{c}{$H_{LBP}+AS$ }		\\
			&	RGB			& 	MI			& 	RGB			& 	MI			& 	RGB   		& 	MI				\\
\hline					
$Acc_{Spx}$		&	63\% (17\%)	&	77\% (13\%)	&	76\% (39\%)	&	88\% (18\%)	&	81\% (20\%)	&	 90\% (6\%)  	\\
\hline
\end{tabular}
\end{adjustbox}
\end{table*}

%% file: tables_images/spAccThConf.tex

\begin{figure}[tbp]
\centering
        \begin{subfigure}[b]{0.23\textwidth}
                \includegraphics[width=\linewidth]{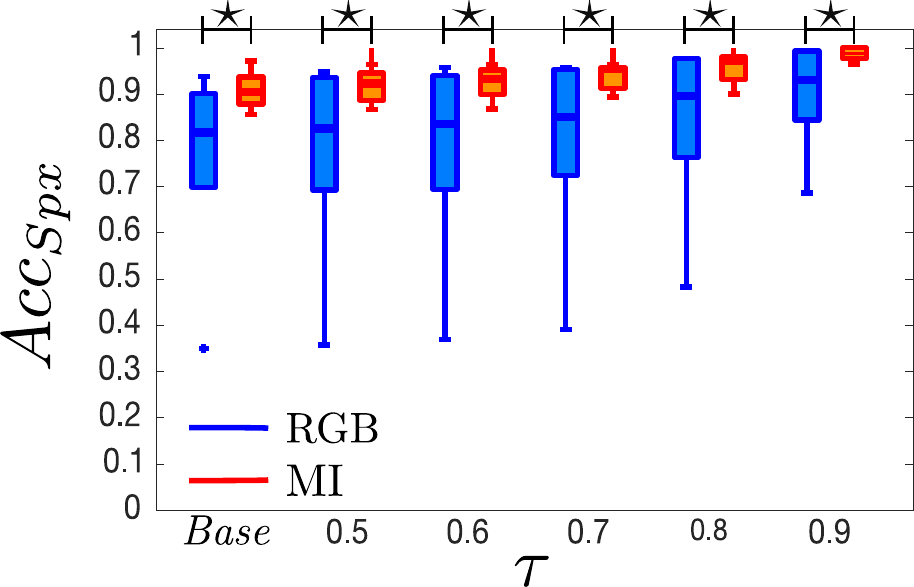}
                \caption{$GC$-based confidence estimation}
                \label{fig:spAccThConf_gc}
        \end{subfigure}%
        ~~~~
        \begin{subfigure}[b]{0.23\textwidth}
                \includegraphics[width=\linewidth]{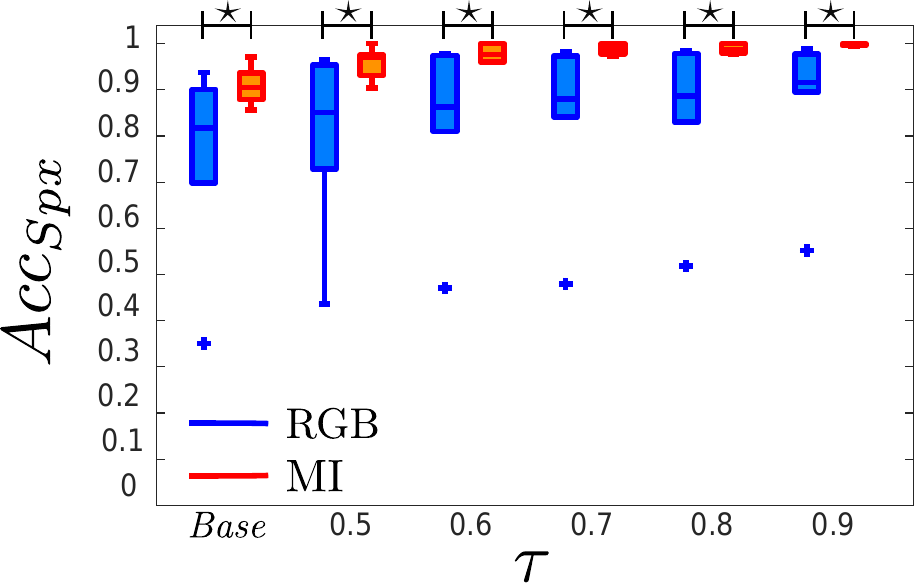}
                \caption{$PPCI$-based confidence estimation}
                \label{fig:spAccThConf_entro}
        \end{subfigure}%
         \caption{Effect of confidence threshold ($\tau$) on the superpixel-based organ classification accuracy rate ($Acc_{Spx}$) for RGB and multispectral imaging (MI).
\textit{Base} refers to classification without confidence estimation. The stars indicate significant differences. The confidence is computed with: (\subref{fig:spAccThConf_gc}) the Gini coefficient ($GC$), (\subref{fig:spAccThConf_entro}) the posterior probability certainty index ($PPCI$).}\label{fig:spAccThConf}
\end{figure}

%% file: tables_images/cm.tex
\begin{figure}[tbp]
    \centering
    \includegraphics[width=1\linewidth]{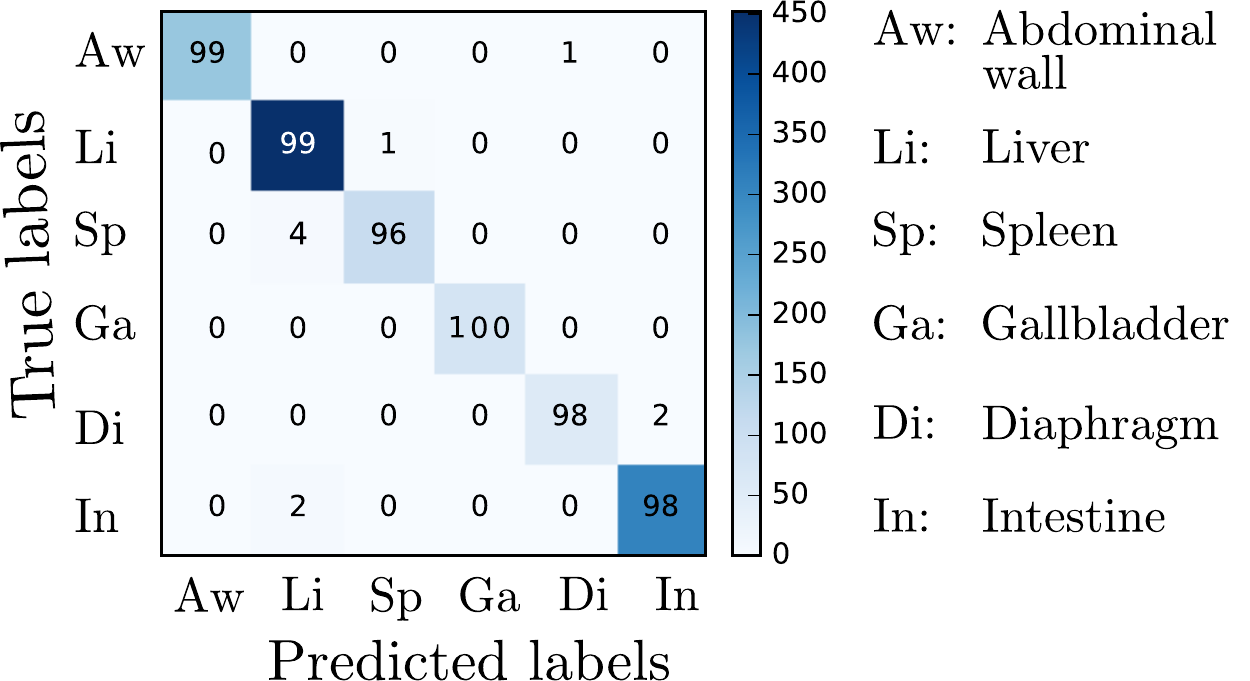}
\caption{Confusion matrix for confidence threshold $\tau=0.9$ on the Gini coefficient and multispectral imaging. The values are in percentages and the colorbar indicates the number of~superpixels.}
\label{fig:cm}
\end{figure}

%% file: tables_images/hl_acc1.tex
\begin{figure}[tbp]
    \centering
\includegraphics[width=.8\linewidth]{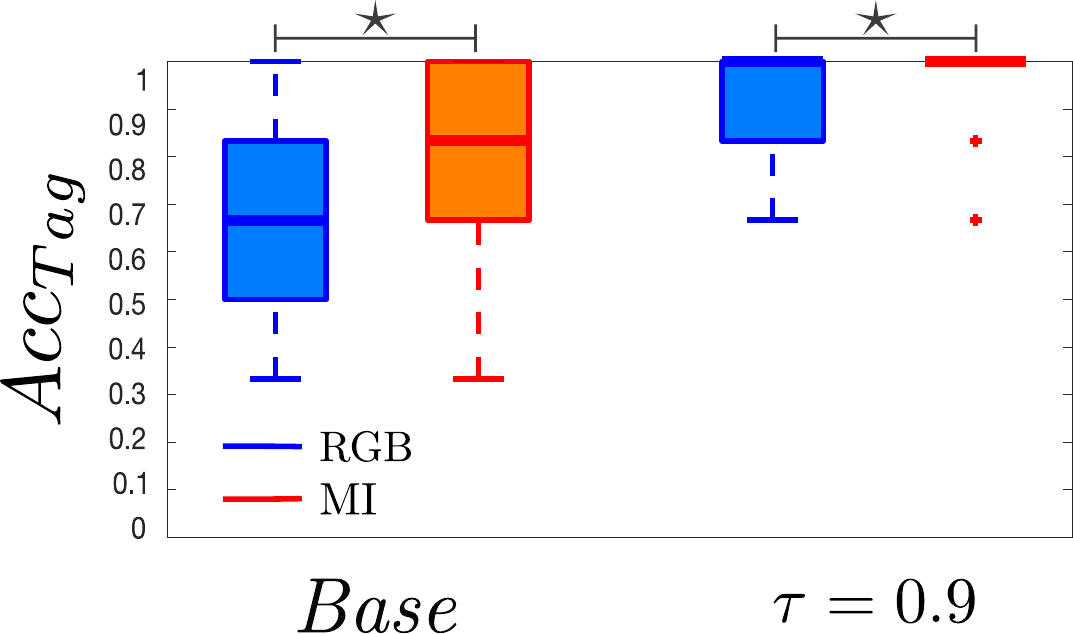}
\caption{Image tagging accuracy ($Acc_{Tag}$) for RGB and multispectral imaging (MI) for \textit{Base} case and following introduction of confidence measure ($\tau=0.9$ on the Gini coefficient). The stars indicate significant differences.}
\label{fig:hl_acc1}
\end{figure}

%% file: tables_images/tag_base_conf.tex
\begin{figure}[tbp]
    \centering
\includegraphics[width=\linewidth]{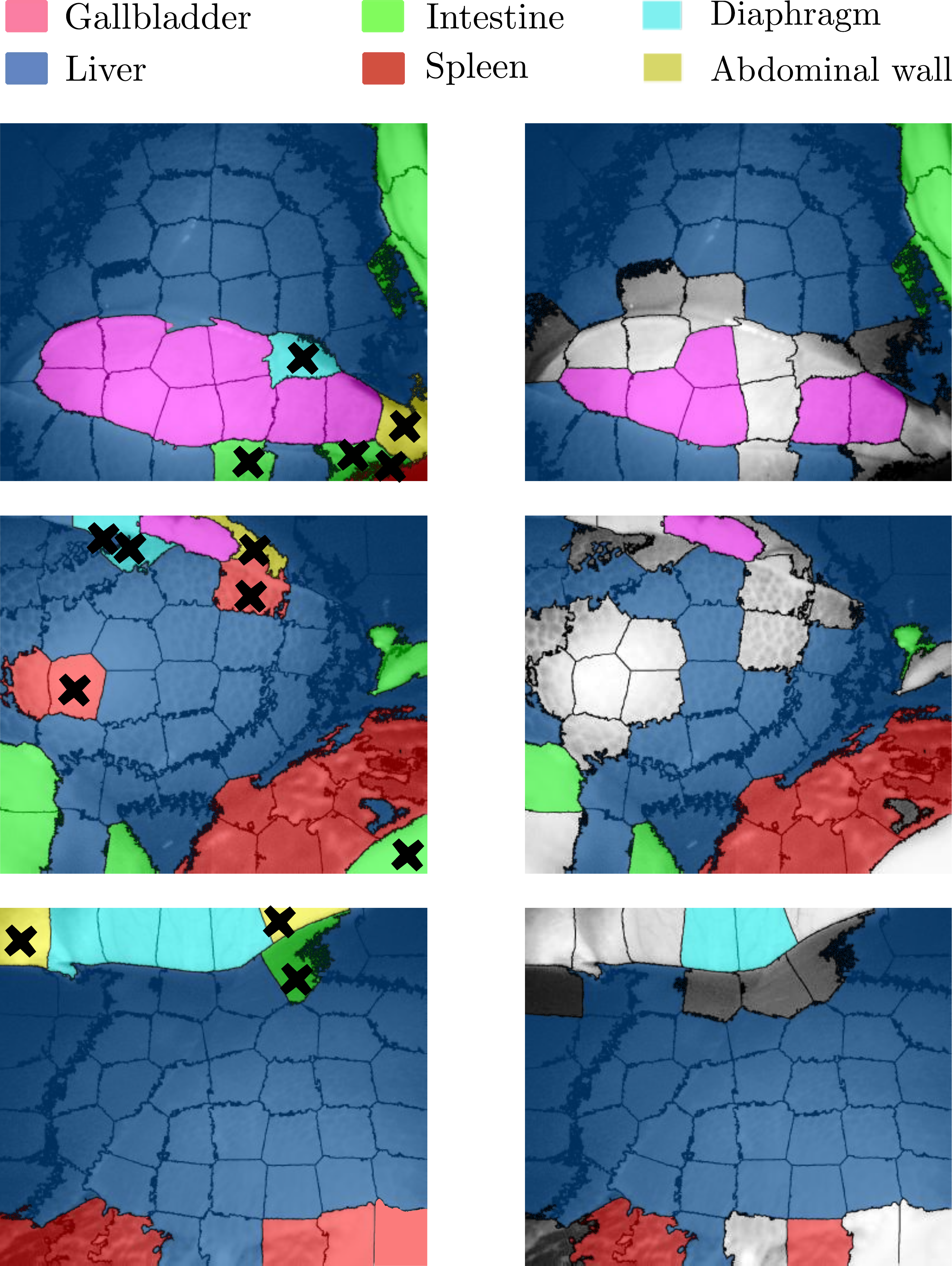}
\caption{{Image tagging examples for \textit{Base} case (left) and  following introduction of confidence measure with $\tau=0.9$ on the Gini coefficient (right). The low-confidence superpixels (in gray) are excluded from the image tagging. The crosses indicate erroneously classified superpixels.}}
\label{fig:tag_base_conf}
\end{figure}

%% file: tables_images/visual.tex
\begin{figure*}[tbp]
\captionsetup[subfigure]{position=b}
\centering
\subcaptionbox{Test image \label{fig:a}}{\includegraphics[width=.23\linewidth]{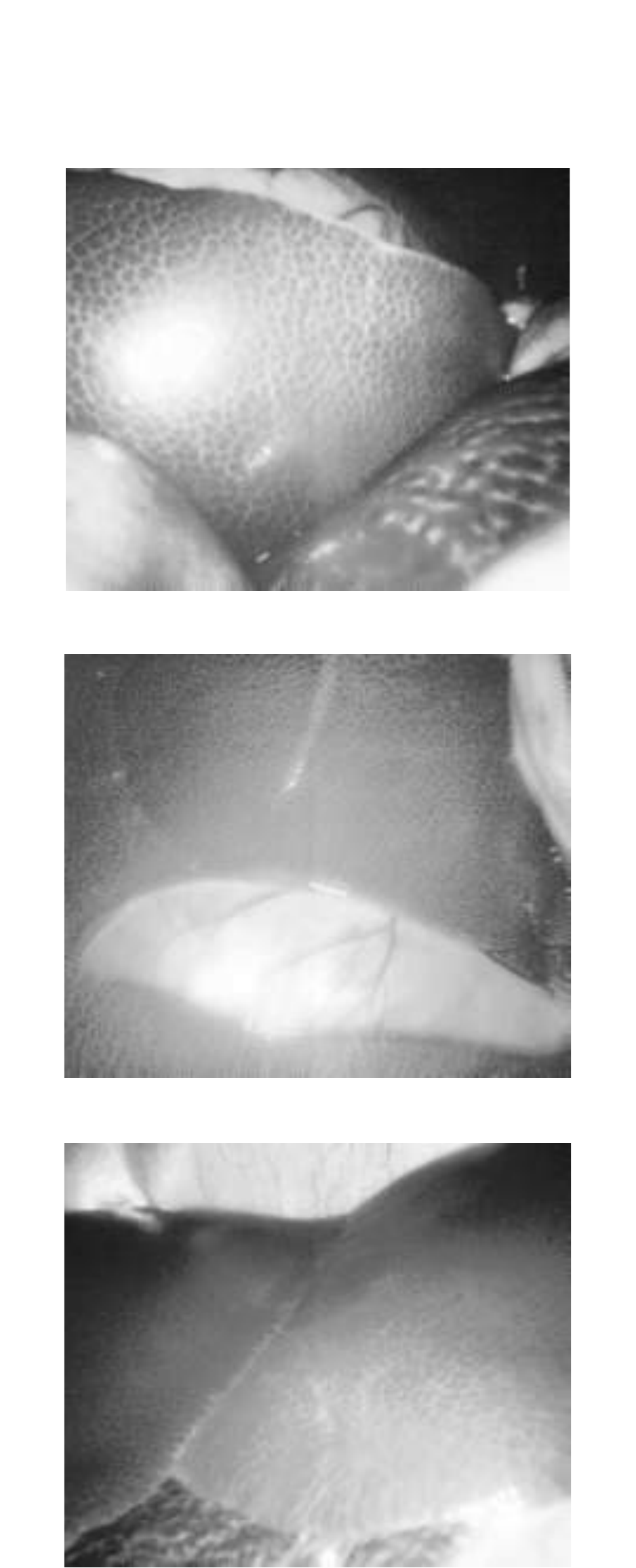}}
~
\subcaptionbox{Superpixel segmentation \label{fig:b}}{\includegraphics[width=.23\linewidth]{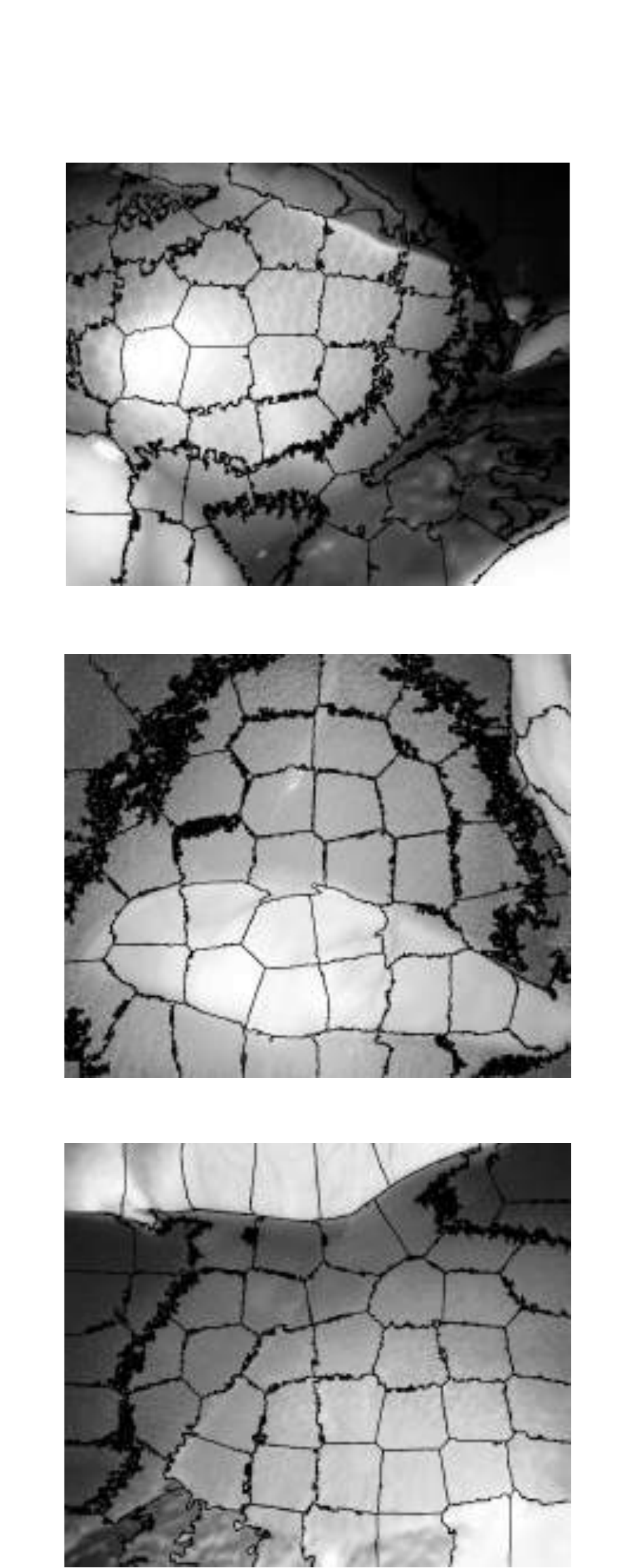}}
~
\subcaptionbox{Classification for confident superpixels\label{fig:c}}{\includegraphics[width=.23\linewidth]{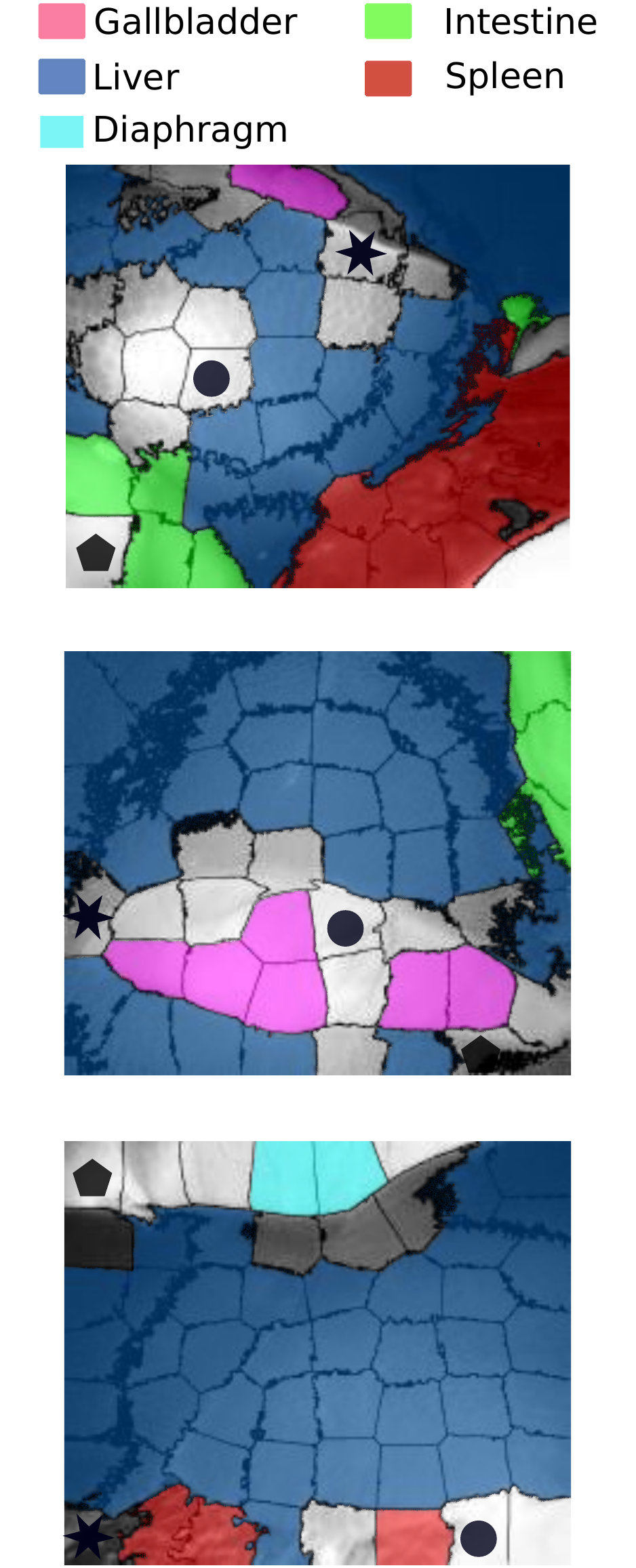}}
~~~
\subcaptionbox{Confidence map with associated colorbar \label{fig:d}}{\includegraphics[width=.23\linewidth]{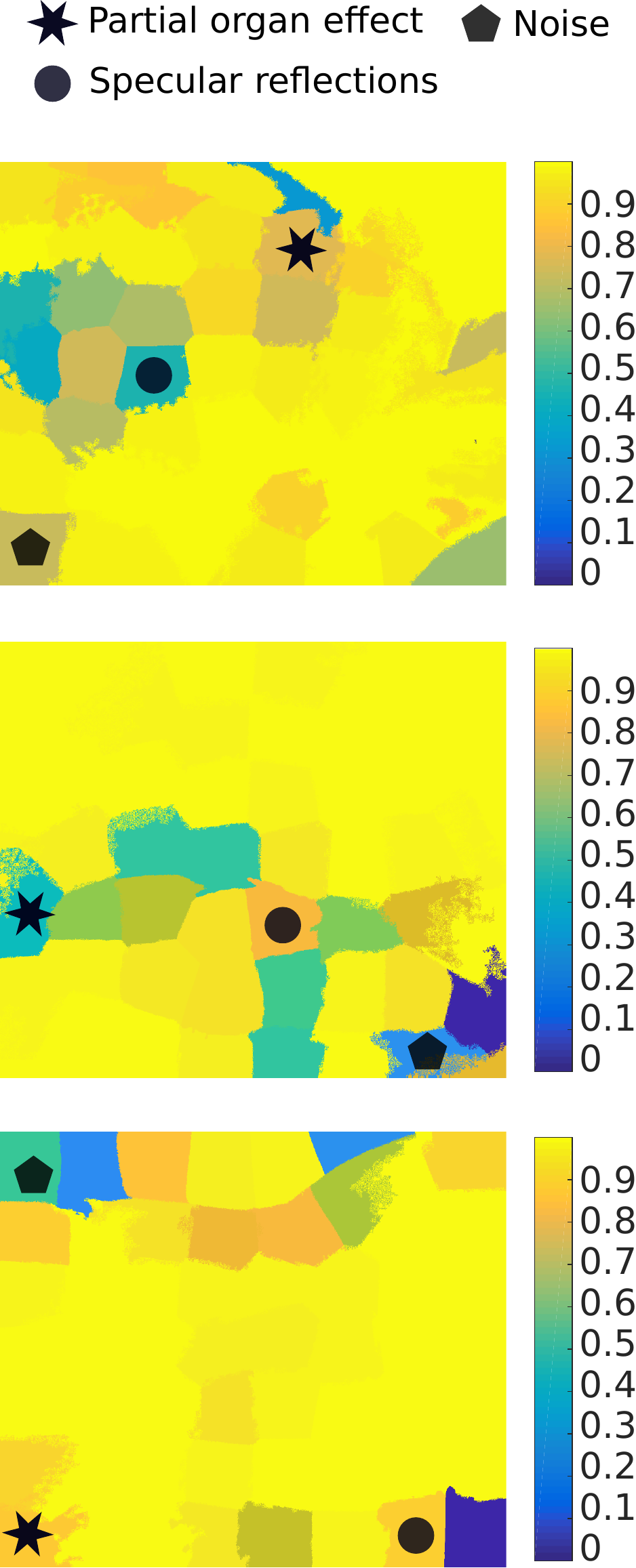}}
\caption{{(a) Test image, (b) test image with superpixel segmentation, (b) corresponding classification for superpixels with acceptable confidence level and (c) confidence map obtained with confidence threshold $\tau = 0.9$ on the Gini coefficient. The symbols give examples of the probable causes of uncertainty.}}
\label{fig:visual}
\end{figure*}